\documentclass[11pt]{article}

\usepackage[preprint]{acl}

\usepackage{times}
\usepackage{latexsym}
\usepackage{amsmath}
\usepackage{graphicx}
\usepackage{multirow}
\usepackage{threeparttable}

\usepackage{rotating}
\usepackage{array}
\usepackage{booktabs}
\usepackage{placeins}
\usepackage{float}
\usepackage{afterpage}

\usepackage[T1]{fontenc}

\usepackage[utf8]{inputenc}

\usepackage{microtype}

\usepackage{inconsolata}

\usepackage{graphicx}

\usepackage{makecell}

\usepackage{CJKutf8}
\usepackage{color}

\usepackage{framed}
\usepackage{tcolorbox}
\tcbuselibrary{breakable}

\newif\ifprint
\printtrue

\title{A Survey on Transformer Context Extension: \\ Approaches and Evaluation}

\author{
 \textbf{Yijun Liu\textsuperscript{1}},
 \textbf{Jinzheng Yu\textsuperscript{2}},
 \textbf{Yang Xu\textsuperscript{1}},
 \textbf{Zhongyang Li\textsuperscript{3}},
 \textbf{Qingfu Zhu\textsuperscript{1}}
\\
 \textbf{\textsuperscript{1}} Research Center for Social Computing and Interactive Robotics \\ \textbf{\textsuperscript{1}} Harbin Institute of Technology
 \\
 \textbf{\textsuperscript{2}} State Key Laboratory of Media Convergence and Communication \\ \textbf{\textsuperscript{2}} Communication University of China
 \\
 \textbf{\textsuperscript{3}} Huawei Technologies, Co., Ltd.
 \\
 \texttt{\{yijunliu, qfzhu\}@ir.hit.edu.cn}
 \\
}

\begin{document}
\maketitle
\begin{abstract}

Large language models (LLMs) based on Transformer have been widely applied in the filed of natural language processing (NLP), demonstrating strong performance, particularly in handling short text tasks.
However, when it comes to long context scenarios, the performance of LLMs degrades due to some challenges.
To alleviate this phenomenon, there is a number of work proposed recently.
In this survey, we first list the challenges of applying pre-trained LLMs to process long contexts.
Then systematically review the approaches related to long context and propose our taxonomy categorizing them into four main types: positional encoding, context compression, retrieval augmented, and attention pattern. 
In addition to the approaches, we focus on the evaluation of long context, organizing relevant data, tasks, and metrics based on existing long context benchmarks. 
Finally, we summarize unresolved issues in the long context domain and put forward our views on future developments.

\end{abstract}

\section{Introduction}

In recent years, the Transformer~\citep{vaswani2017attention} architecture has made significant progress in many NLP tasks~\citep{devlin2018bert, radford2018improving, lewis2019bart, raffel2020exploring, brown2020language, chen2021evaluating, cobbe2021training}, and has become the foundational model of many applications. 
Large language models can handle tasks involving short texts, within the pre-trained context length. 
However, current scenarios always involve longer texts, such as book-/repo- level tasks~\citep{sharma2019bigpatent, liu2023neural, zhang2023repocoder, liu2023repobench}, dialogue systems with long contexts~\citep{dey2022dialogen, li2024streamingdialogue}, content-rich in-context learning~\citep{li2024long} and so on. 
The performance of the pre-trained LLMs degrades and the models often fail to utilize the complete knowledge contained within the long context inputs.
This may be caused by three inherent challenges: out-of-distribution (OOD) problem~\citep{han2024lm}, "Lost in the Middle" phenomenon~\citep{liu2024lost} , and quadratic complexity of attention~\citep{zhou2024survey}.
Recently, a lot of work has been proposed to improve and evaluate models' ability to handle long contexts in the community.

\begin{figure*}[t]
  \centering
  \includegraphics[width=\linewidth]{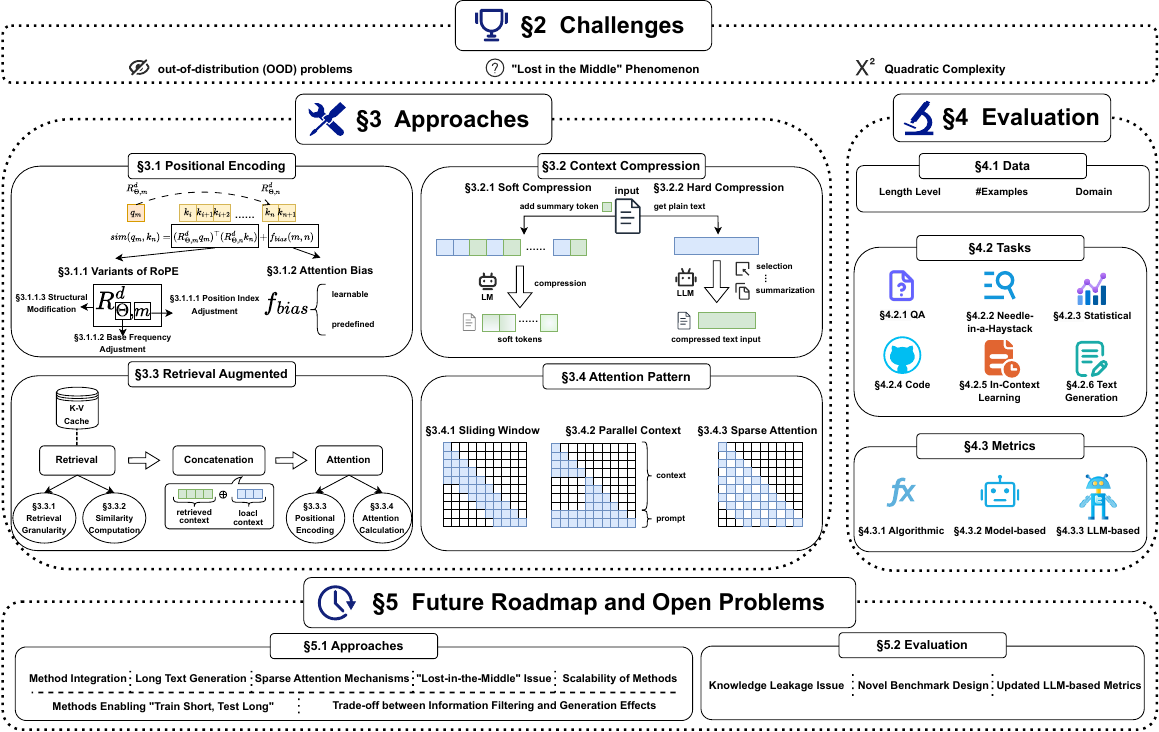} 
  \caption{Framework of survey. 
  We first list three inherent challenges in Section~\ref{sec:challenges}. 
  And then we systematically review related approaches and propose a novel taxonomy with four major categories in Section~\ref{sec:approaches}. 
  Next, in Section~\ref{sec:evaluation}, we organize the evaluation aspect from three perspectives: data, tasks, and metrics based on existing benchmarks.
  At last, we show our views on future roadmap and open problems in Section~\ref{sec:future}.
  }
  \label{fig:framework}
\end{figure*}

This survey focuses on approaches and evaluation in the long context field, systematically reviewing existing related work.
As illustrated in Figure~\ref{fig:framework}, we propose a novel taxonomy for approaches, categorizing them into four main groups: positional encoding, context compression, retrieval augmented, and attention pattern. 
Additionally, we focus on the evaluation aspect and organize work on data, tasks, and metrics based on existing benchmarks. 
In addition to the two main parts of approaches and evaluation, we present our viewpoints on the current unsolved issues and potential future directions in the long context domain.
To illustrate the current status more theoretically, we also list the main challenges in the field of long context before introducing specific work.
Although most existing methods and benchmarks have not corresponded to them, these challenges are still instructive for the development of approaches and evaluation.

There are also some surveys that focus on the long context domain.
They each have their own emphasis, but there is no systematic and comprehensive survey of approaches and evaluation in the field of long context, which can provide researchers with a quick and efficient guide.
Some surveys only include a part of the methods, lacking a comprehensive overview of the approaches related to long context.
\citet{zhao2023length} focus on work addressing length extrapolation from the perspective of positional encoding, while there are some surveys from the perspective of KV Cache~\citep{li2025surveylargelanguagemodel, shi2024costdownreviewmethods}.
Besides, though some surveys have categorized existing work, their taxonomies are not clear, and there are overlaps between categories.
For example,~\citet{huang2023advancing} divide the methods for enhancing Transformer architecture models into five categories, but some existing methods can belong to multiple categories.
And~\citep{pawar2024and} also has this problem, which distinguishes existing techniques into two categories: interpolation and extrapolation.
Also, some surveys even involve some common methods that not specifically designed for long contexts.
\citet{dong2023survey} provide an overview of the text-preprocessing methods, architectures, special characteristics and application for long context, but they cover some general topics.
What's more, these surveys pay little or even no attention to the evaluation aspect.

To fill the above gap, our survey proposes a novel and comprehensive taxonomy on both approaches and evaluation aspects.
It is worth noting that we focus on work that applies Transformer-based models to long text tasks, but not improving Transformers (nor other architectures) in a universal scenario.
That is to say, this survey does not cover fields like long chain-of-thought reasoning~\citep{chen2025reasoningerasurveylong}, multimodal long context~\citep{song2024milebench, qiu2024longhalqalongcontexthallucinationevaluation}, efficient Transformer~\citep{zhou2024survey}, and State Space Model (SSM)~\citep{wang2024statespacemodelnewgeneration}.
In addition, the long context we focus on is the long input content, rather than the introduction of external knowledge in the Retrieval-Augmented Generation (RAG) scenario \citep{yu2024evaluationretrievalaugmentedgenerationsurvey, zhao2024retrievalaugmentedgenerationaigeneratedcontent, fan2024surveyragmeetingllms}.

\section{Challenges}
\label{sec:challenges}

When applying pre-trained LLMs to the long context scenarios, there are some inherent challenges affecting models' performance.
We list the three most important and common challenges: OOD problem, "Lost in the Middle" phenomenon, and quadratic complexity.

\paragraph{OOD Problem}When processing sequences that exceed pre-trained context window length, the models face out-of-distribution (OOD) problems.
\citet{han2024lm} verify theoretically and empirically that three key factors contribute to OOD issues, thereby limiting models' extrapolation capabilities: 1) unseen inter-token distances, 2) increased number of attended tokens, and 3) implicitly encoded position information of the starting tokens.

\paragraph{"Lost in the Middle" Phenomenon}\citet{liu2024lost} discover the "Lost in the middle" phenomenon through experiments that when LLMs receive a long input, they tend to focus on the information at both the beginning and end of the input sequence.
At the same time, they neglect the content in the middle, thus failing to capitalize on the key information within the long input.

\paragraph{Quadratic Complexity} Due to the quadratic complexity of attention, directly using pre-trained LLMs for training or inference on long context is time and resource consuming~\citep{zhou2024survey}.

The above are three inherent challenges in the field of long context, and some existing methods have alleviated them to a certain extent.
But it is worth noting that most of the methods do not start from this perspective. 
They consider directly improving the performance of downstream tasks.
However, we believe that these three challenges are still the fundamental problems that need to be solved. 
They play a vital role in the proposal of methods and construction of benchmarks.
Moreover, they are the focus of subsequent research.

\section{Approaches}\label{sec:approaches}
For the existing approaches for long context, we summarize their characteristics and propose a novel taxonomy different from previous work. As illustrated in Figure~\ref{fig:framework}, mainstream methods are divided into four major categories: positional encoding, context compression, retrieval augmented, and attention pattern, which are introduced below. For more details, please see the Appendix~\ref{appendix:approaches}.
\subsection{Positional Encoding}
\citet{kazemnejad2024impact} mention that positional encoding (PE) appears to be a major factor in the length generalization of Transformer.
During the inference process, when encountering sequences that exceed the length of the pre-trained window, the model needs to handle the position index that was not encountered during pre-training.
This may lead to Out-Of-Distribution (OOD) issues. 
Thus, we would like to find an appropriate positional encoding method that allows the model to effectively encode position in sequences that exceed the pre-trained window length.
Based on the implementation methods, solutions can be categorized into two main types: Variants of Rotary Position Embedding (RoPE,~\citealp{su2024roformer}) and Attention bias method.
The following sections will detail these two methods.

Though the designed positional encoding strategies can alleviate the extrapolation problem, experiments have found that models without positional encoding (NoPE) show better performance than these methods in reasoning tasks~\citep{kazemnejad2024impact} .
That's because when causal masks are used for decoding, the model reads the sequence sequentially from left to right.
And this process naturally incorporates the sequential information of the token. 
This finding suggests that when designing a position encoding strategy, we may need to consider the way the model processes sequences and the requirements of the task.

\subsubsection{Variants of RoPE}
Rotary Position Embedding (RoPE, ~\citealp{su2024roformer}) is a positional encoding method utilized in a series of models such as~\citet{gpt-j, touvron2023llama, roziere2023code}.
RoPE incorporates explicit relative position dependency in self-attention, which can be expressed as
\begin{equation}
\begin{aligned}
\operatorname{sim}(q_m,k_n) & = q_m^\top R^d_{\Theta, n-m}k_n \\
& = (R^d_{\Theta, m}q_m)^\top(R^d_{\Theta, n}k_n) \text{\, ,} 
\end{aligned}
\end{equation}
where ${R^d_{\Theta,m}}$ is called the rotation matrix. 
The original RoPE's extrapolation capability is not very robust and can only maintain performance slightly beyond the pre-trained context length. 
Consequently, existing work enhances RoPE for better extrapolation.
The core of RoPE is the rotation matrix ${R^d_{\Theta,m}}$, which is parameterized by the position index $m$ and the function family $\Theta$.
We can optimize RoPE by adjusting these parameters or even the structure of RoPE itself.
Existing related work can be divided into three subcategories: position index adjustment, base frequency adjustment, and structural modification.
\paragraph{Position Index Adjustment}
This method involves modifying the allocation or calculation of position index $m$ to maintain the relative distances between tokens within the model's pre-trained index range.
This can be implemented in three ways.
We can adjust the allocation of the position index $m$ \citep{an2024training}.
Besides, proportionally scale $m$ for long sequences to adapt to the pre-trained window \citep{chen2023extending}. 
What's more, we can combine the above two methods, reallocating position index to some tokens in the sequence, while scaling the position index for others \citep{rerope2023}. 
\paragraph{Base Frequency Adjustment}
From the formula of rotation matrix (see details in Appendix~\ref{appendix:rope}), we can see that each non-zero term is a trigonometric function value with $\theta_i$ as independent variable.
And the value of $\theta_i$ affects the effect of rotation matrix to a certain extent.
Base frequency adjustment is to enhance the model extrapolation performance by modifying $\theta_i$ in the trigonometric function terms in the rotation matrix.
NTK (Neural Tangent Kernel) theory shows that when the input dimension is low and its embedding representation lacks high-frequency components, it is difficult for the neural network to learn high-frequency information ~\citep{tancik2020fourier}.
Therefore, researchers choose to adjust $\theta_i$ with the idea of ``extrapolation on high-frequency and interpolation on low-frequency''.
One strategy is to change the base $b$ of the exponential terms $\theta_i$ in the function cluster $\Theta$, and change it from the default value $b = 10000$ to other values which can improve the model extrapolation performance \citep{peng2023ntk, roziere2023code}. 
Another strategy is to directly scale $\theta_i$ \citep{blocntkparts, peng2023yarn}. 

\paragraph{Structural Modification}
The methods described above focus on modifying variables in RoPE formula without altering its basic structure.
Some existing work explores adjusting the structure of RoPE itself to better extrapolate, which optimizes the original RoPE formula~\citep{sun2022length}. 

\subsubsection{Attention Bias}
This type of method introduces relative position information by adding a bias related to the relative distance between tokens when calculating the similarity between query and key vectors.
The process can be expressed as follows: 
\begin{equation}
\begin{aligned}
\operatorname{sim}{(q_m, k_n)} = {q_m^{\top}k_n + f_{bias}(m,n) \text{\, ,} }
\end{aligned}
\end{equation}
where $f_{bias}(m,n)$ is a bias function that depends on the token position index corresponding to query and key. $f_{bias}(m,n)$ be divided into two categories: learnable \citep{raffel2020exploring, chi2022kerple}, predefined \citep{press2021train, chi2022dissecting}.

\subsection{Context Compression}
Existing work proposes compressing the long input sequence into a shorter one for representation.
These methods can be categorized into two main types by the compression granularity: soft compression and hard compression. 
\subsubsection{Soft Compression}
In order to shorten the sequence length, the soft compression method uses the model to compress the original input token sequence into a shorter summary token sequence.
These summary tokens are soft tokens which act as compression representation but do not correspond to words with actual meaning.
They are inserted into the original token sequence to form a new input.
During the forward pass of the model, the information from the original token sequence is gathered into the summary token sequence, which represents the original input for subsequent operations.
Since summary tokens do not appear during the model's pre-training, additional training is necessary for the model to learn how to generate and utilize these tokens \citep{bulatov2022recurrent, li2023recurrent, chevalier2023adapting, ge2023context, mu2024learning}. 
This method can shorten the length of the hidden vector sequence, so that enabling it to be processed within the model's pre-trained window.

\subsubsection{Hard Compression}
This method utilizes some techniques to directly shorten plain text sequence length.
This process can be achieved through selection and summarization. 
It doesn't introduce additional tokens and targeted training, which makes it can be applied to some black box models \citep{jiang2023llmlingua, jiang2024longllmlinguaacceleratingenhancingllms, chen2023walking}.

\subsection{Retrieval Augmented}
Some existing work propose retrieval-augmented methods to enhance model performance on long context tasks by selectively incorporating crucial tokens from history context into attention.
With reference to related work, we summarize a processing paradigm for this type of method.
Initially, the \textit{(key, value)} pairs from history are stored in the KV cache.
Then the model retrieves the corresponding token representations from the KV cache at different retrieval granularity levels.
This process is based on the similarity between current token and history tokens from KV cache. 
The top-k relevant tokens are selected as the retrieved context, which is then concatenated with the context within the current window to form a new input.
Finally, the model applies appropriate positional encoding to this concatenated context for attention calculation.
Below, we summarize the different methods according to each step of the above process:

\subsubsection{Retrieval Granularity}
In the process of long context retrieval, we focus on the most relevant subset of tokens from KV cache related to the current processing step. 
Different methods use different retrieval granularity, with the basic being token-level retrieval.
Specifically, it involves calculating the similarity of each history token in the KV cache with the current token, and selecting the top-k history tokens' key and value vectors as the retrieval result. 
Methods applying this strategy include MemTRM~\citep{wu2022memorizing}, FoT~\citep{tworkowski2024focused}, Unlimiformer~\citep{bertsch2024unlimiformer}, etc. 
Besides, some work focus on block-level retrieval, which retrieve top-k set of tokens in one step~\citep{wang2024augmenting, rubin2023long, xiao2024infllm, mohtashami2024random} .

\subsubsection{Similarity Computation}

Almost all existing works compute the inner product of query and key as similarity. 
This strategy draws from the standard attention mechanism, which calculates the dot product between the query and key to allocate corresponding weights to the value~\citep{vaswani2023attentionneed}. 
It is simple to implement and can effectively capture and utilize the similarity information between queries and keys.

\subsubsection{Positional Encoding}

After computing the similarity, we select the top-k relative tokens as the results, and call them retrieved context tokens. 
Correspondingly, tokens within the current window are called as local context tokens. 
These two types of context tokens are concatenated to form a new set of context tokens. 
Before these new context tokens are fed into the model for attention computation, it is necessary to consider suitable positional encoding to distinguish the information of tokens at different positions.
Some work choose to assign the same position vector to the retrieved context tokens~\citep{wu2022memorizing, tworkowski2024focused, xiao2024infllm}, while~\citet{mohtashami2023landmark} choose reallocation strategies.

\subsubsection{Attention Calculation}
Next, when performing attention calculation, we need to consider how to make full use of retrieved context tokens and local context tokens.
Different approaches use different strategies. 
Simply, ~\citet{tworkowski2024focused, xiao2024infllm}choose standard attention, while~\citet{bertsch2024unlimiformer} chooses cross attention. Besides,~\citet{wu2022memorizing, wang2024augmenting} adopt a Joint Attention method. Landmark employs the Grouped Softmax method, a fine-grained approach for calculation~\citep{mohtashami2023landmark}.

\subsection{Attention Pattern}
There is a class of methods modifying the attention pattern, i.e. the range of tokens attended to. 
They can better adapt models to expand processing sequence length.
Some of them do not require additional training and can be employed as plug-and-play solutions in existing models. 
These methods can be divided into three main categories: sliding window, parallel context, and sparse attention.

\subsubsection{Sliding Window}

This type of method divides the sequence into segments and performs attention calculation segment by segment without significantly increasing computational complexity. 
The attention results from earlier segments are stored, which later segments can use during their attention calculation~\citep{dai2019transformer, han2024lm, xiao2023efficient}. 

\subsubsection{Parallel Context}

The Parallel Context method folds the context part of the input (e.g., in-context examples) into multiple segments. 
These segments first calculate attention independently, and share the same set of position indexes.
And then prompt tokens in the input attend to all the context tokens, so that fully utilize contextual information ~\citep{ratner2022parallel, hao2022structured}.
These methods require no training and can be plug-and-played into existing models.

\subsubsection{Sparse Attention}

Some work reduce the number of tokens involved in the attention computation, decreasing computational load.
They abandon the original attention method which attends to local continuous tokens, while expand the attentive field, and attend to discrete tokens from further context ~\citep{ding2023longnet, yu2023megabyte, chen2023longlora}.

\section{Evaluation}\label{sec:evaluation}

\begin{table*}
    \centering
    \small
    \renewcommand{\arraystretch}{1.2} 
    \resizebox{\textwidth}{!}{
    \begin{tabular}{l|lrl}
        \hline
        \makecell[c]{\textbf{Benchmark}} & \makecell[c]{\textbf{Length Level}} & \makecell[c]{\textbf{\#Examples}} & \makecell[c]{\textbf{Domain}}\\
        \hline
        
        \multirow{1}{*}{SCROLLS~\citep{shaham2022scrolls}} & \multirow{1}{*}{1k$\sim$4k}  &119,495& Literature, Dialog \\
       
        \multirow{1}{*}{ZeroSCROLLS~\citep{shaham2023zeroscrolls}} & \multirow{1}{*}{0k$\sim$16k}  &4,378 &Wiki, Literature, Dialog \\
        
        \multirow{1}{*}{LongBench~\citep{bai2023longbench}} & \multirow{1}{*}{0k$\sim$4k, 4k$\sim$8k, $>$8k}  &\multirow{1}{*}{4,750}&Wiki, Literature, Dialog, Report, Code, News\\
       
        \multirow{1}{*}{LooGLE~\citep{li2023loogle}} & \multirow{1}{*}{0k$\sim$24k}  &776 & Wiki, Paper\\
        
        \multirow{1}{*}{BAMBOO~\citep{dong2023bamboo}} & \multirow{1}{*}{0k$\sim$4k, 4k$\sim$16k}  &1,502 & Wiki, Dialog, Report, Code, Paper\\

        \multirow{1}{*}{LongICLBench~\citep{li2024long}} & \multirow{1}{*}{2k$\sim$50k}  & 3,000 & Dialog, News, Common Sense\\
        \hline

        \multirow{1}{*}{L-Eval~\citep{an2023eval}} & \multirow{1}{*}{3k$\sim$200k}  &\multirow{1}{*}{411} & Literature, Dialog, News, Paper, Common Sense\\
        
        \multirow{1}{*}{Ada-LEval~\citep{wang2024ada}} & \multirow{1}{*}{1k$\sim$128k}  & 117,500 & Literature, Code\\

        \multirow{1}{*}{$\infty$Bench~\citep{zhang2024bench}} & \multirow{1}{*}{0k$\sim$200k}  & 3,946 & Literature, Dialog, Code\\

        \multirow{1}{*}{NeedleBench~\citep{li2024needlebench}} & \multirow{1}{*}{1k$\sim$4k/8k/32k/128k/200k/1m$+$}  & \multirow{1}{*}{-} & Wiki, Literature, Dialog, Report, Code, News   \\
        
        \multirow{1}{*}{LV-Eval~\citep{yuan2024lv}} & \multirow{1}{*}{0k$\sim$16k/32k/64k/128k/256k}  & \multirow{1}{*}{1,729} & Wiki, Literature, Dialog, Report, Code, News, Paper  \\
        \hline
    \end{tabular}}
    \caption{Statistics on data characteristics of the datasets in existing long context benchmarks. 
    Length level represents the range of token lengths in the dataset used in the benchmark. 
    \#Examples refers to the total number of examples. 
    Domain denotes the data sources. 
    The corresponding contents in table are directly extracted or calculated from the original papers.
    Given that current models mainly within context windows exceeding 100k tokens, we categorize benchmarks based on this threshold. 
    Benchmarks with contexts exceeding 100K tokens are listed in the lower part.
    }
    \label{tatistics of benchmarks}
\end{table*}

In the long context scenario, evaluating the model's ability to understand and utilize long context is also a new and critical issue.
But as described before, current surveys pay little or even no attention to the evaluation aspect.
To fill this gap, we summarize the data, tasks, and metrics of long context evaluation in our survey based on existing benchmarks.
The following is a brief introduction, detailed information is in the Appendix~\ref{appendix:evaluation}.

\subsection{Data}

In order to explore what data should be used to test model's ability to process long context, we conduct a statistical analysis of datasets in existing benchmarks and summarize their data characteristics.

The evaluation of a model's long context capability requires not only the long data but also the data diversity and quality.
As shown in Table~\ref{tatistics of benchmarks}, we focus on three characteristics of the datasets in existing long context benchmarks: length level, total number of examples, and the domain it covers.

Besides, we also discuss about knowledge leakage issue, which need to be addressed when constructing the dataset, in the Appendix~\ref{appendix:knowledgeleakageissue}

\subsection{Tasks}

Currently, existing benchmarks propose numerous tasks to evaluate the model's ability to process long context. 
But there is no systematic taxonomy for these tasks.
Therefore, we divide all tasks in existing benchmarks into seven categories from the perspective of task setting : Question Answering, Needle-in-a-Haystack, Statistical Tasks, Code, In-Context Learning, Text Generation and Other Tasks.
Below is the introduction of each type of task, and the details are in the Appendix~\ref{appendix:tasks}.

\subsubsection{Question Answering}
\textbf{Single-hop Question Answering} requires models to locate and extract answers from a single text passage, typically involving a single fact~\citep{rajpurkar2016squad,joshi2017triviaqa,kovcisky2018narrativeqa}. 

\noindent\textbf{Multi-hop Question Answering} requires models to integrate information from multiple sources to answer complex questions. This often involves reasoning across different pieces of evidence~\citep{ho2020constructing,trivedi2022musique,yang2018hotpotqa,chen2024essential,Zhuang2023ThroughTL}. 

\subsubsection{Needle-in-a-Haystack}

Needle-in-a-Haystack evaluate LLMs' ability to extract specific content from long contexts.
These tasks can evaluate the model's retrieval capability, also measure the range of context lengths model can handle~\citep{zhu2024longembed,mohtashami2023landmark,zhang2024bench,li2024needlebench}.

\subsubsection{Statistical Tasks}
\textbf{Long Arithmetic Calculation} requires models to perform addition and subtraction operations on lengthy arithmetic expressions~\citep{zhang2024bench,zhang2023marathon,cobbe2021training,xu2024can,Chen2024UnlockingTB}.

\noindent\textbf{Numerical Information Extraction} requires models to identify specific mathematical elements~\citep{zhang2024bench,li2023loogle}.

\noindent\textbf{Sentiment Aggregation } requires models to output the percentage of positive reviews when provided with a collection of reviews~\citep{angelidis2021extractive,shaham2023zeroscrolls}.

\noindent\textbf{Paragraph Counting} requires models to count the number of unique paragraphs in a set of randomly repeated and shuffled passages~\citep{bai2023longbench}.

\noindent\subsubsection{Code}

\noindent\textbf{Code Completion} requires models to complete missing code fragments based on preceding code and context~\citep{chen2021evaluating,zheng2023codegeex,bai2023longbench,guo2023longcoder,zan2022language,dong2023bamboo,qin2024large}. 

\noindent\textbf{Code Running} asks models to infer the output of lengthy programs by tracing a series of cascading function calls~\citep{bubeck2023sparks,an2023eval,zhang2024bench}.

\noindent\textbf{Code Debugging} requires models to identify deliberately inserted errors~\citep{zhang2024bench}.

\subsubsection{In-Context Learning}

The input will contain a certain amount of examples, resulting in a long input. 
This is caused by the example itself is very long or the number of examples is particularly large.
Based on this fact, we divide In-Context Learning task into two categories : long example learning and many-shot learning.

\noindent\textbf{Long Example Learning} requires models to process extensive inputs with long examples which have large label spaces and generate accurate predictions.
This task inherently is a long-context challenge~\citep{bai2023longbench,li2024long,li2002learning,nlpcc14st}.

\noindent\textbf{Many-shot Learning} leverages the expanded context windows of models to process hundreds or even thousands of examples in order to complete a given task~\citep{yu2020dialogue,bertsch2024context}.

\subsubsection{Text Generation}

\noindent\textbf{Language Modeling} serving as the pre-training task for LLMs, is also a widely used basic task to test the model's ability to generate text.

\noindent\textbf{Document Summarization}
requires models to make a summary of the input documents, which encompasses single-document and multi-document tasks.
Single-document summarization extracts key information from a single document~\citep{wang2022squality,chen2021summscreen,huang2021efficient,zhong2021qmsum}, while multi-document summarization synthesizes information from multiple sources into a comprehensive, non-repetitive summary containing all key points~\citep{bai2023longbench,an2023eval,fabbri2019multi}.

\noindent\textbf{Open-ended Text Generation}
requires models to produce coherent and logical content on given topics~\citep{tan2024proxyqa,bai2024longwriter,kumar2024longlamp,ni2024xl,rafailov2024direct}.

\subsubsection{Other Tasks}

In addition to the six types of tasks listed above, there are some tasks that are not included in this classification system but are equally important for testing the model's long context ability.

\noindent\textbf{Reordering} asks models to reconstruct the original sequence of shuffled fragments by considering the broad context and logical relationships~\citep{kryscinski2021booksum, shaham2023zeroscrolls, li2023loogle, dong2023bamboo, wang2024ada}.

\noindent\textbf{Context Consistency} shows models an academic paper and a hypothesis, requiring models to judge whether the hypothesis is supported or contradicted by the ideas in the paper~\citep{dong2023bamboo}. 

\noindent\textbf{Summary Source Paragraph Identification}
challenges models to identify the original source paragraphs for given summaries~\citep{bai2023longbench}.

\noindent\textbf{Character Identification}
requires models to identify different speakers in long dialogues by recognizing their distinct characteristics~\citep{TVMEG,WelshParliamentRecord,zhang2024bench, dong2023bamboo,chen2021summscreen}.

\subsection{Metrics}
In addition to data and tasks, metrics can directly reflect the model's ability to handle long contexts.
With current long context task designs gradually changing from classic NLP tasks to more practical tasks, the requirements for metrics are constantly increasing.
We organize metrics for testing models' capabilities on long context according to the three stages of metrics development: Algorithmic Metrics, Model-based Metrics, and LLM-based Metrics.
From these three metrics stages, it can be seen that the metrics development trend becomes more and more complex and flexible.

\subsubsection{Algorithmic Metrics}
Algorithmic metrics are calculated based on the model output or logits through defined formulas.
Their implementation is very simple and can reflect the effect of language modeling and some downstream tasks to a certain extent.

Perplexity (PPL) is one of the most common algorithmic metrics used in existing long context benchmarks~\citep{beltagy2020longformer,roy2021efficient,press2021train}. 
Meanwhile, some benchmarks employ other algorithmic metrics such as accuracy, f1, and N-gram-based metrics (ROUGE,~\citealp{lin2004rouge} and BLEU,~\citealp{papineni2002bleu}, etc.) to evaluate LLMs on certain downstream tasks ~\citep{shaham2023zeroscrolls,bai2023longbench,kasai2021bidimensional}.

However, these algorithmic metrics have several limitations, such as content quality, syntactic accuracy, and human correlation issues~\citep{reiter2009investigation,stent2005evaluating,sun2021long,an2023eval,improvingnot,tan2024proxyqa}.
This causes algorithmic metrics to be limited in reflecting the model's ability to process long context.
A number of approaches have been developed to improve algorithmic metrics.
Such as enhancing scoring techniques, restructuring task formats and so on~\citep{yuan2024lv,dong2023bamboo,li2024needlebench}.

\subsubsection{Model-based Metrics}

To improve the consistency with human judgments, pre-trained language models are being employed to evaluate~\citep{zhang2020bertscore, yuan2021bartscore}.
Specifically, pre-trained models (such as BERT,~\citealp{devlin2018bert}, BART,~\citealp{lewis2019bart}, etc.) are used to calculate the similarity score between the model output and reference text to evaluate the performance of downstream tasks.

However, these model-based metrics entirely rely on representations learned from pre-trained language models and require reference texts.
They may not be accurate enough for evaluating some novel and creative text generation tasks.

\subsubsection{LLM-based Metrics}

Combining the above two metrics issues, LLM-based metrics are proposed, utilizing sufficient knowledge within LLMs for evaluation.
For example, LLM-based metrics prompt LLMs to offer human-like multi-dimensional assessment ~\citep{wang2023chatgpt,li2023loogle,shen2023large,chiang2023can,zhang2024bench,zheng2024judging,liu2023g,tan2024proxyqa,mu2024staticevaluationdynamicapproach}
and interpretable reasoning~\citep{wang2023large,luo2023chatgpt,wu2023large}.
 
LLM-based metrics fundamentally distinguish from the other two metrics, which behave much more mechanically. 
In addition, they demonstrate enhanced agreement with human evaluations~\citep{wang2023chatgpt,li2023loogle}.
Due to this higher consistency and wider scope of application, LLM-based metrics are gaining increasing attention in long-context evaluation.

\section{Future Roadmap and Open Problems}
\label{sec:future}
Despite the rapid development of long context techniques, numerous challenges remain unresolved.
Looking to future roadmap, we list vital open problems and present our perspectives on the developments.
They are also divided into two parts: approaches and evaluation.

\subsection{Approaches}

\noindent\textbf{Method Integration} would combine methods' strengths to address the challenges of extrapolating long context from multiple perspectives.

\noindent\textbf{Long Text Generation} remains under-researched, which concentrate on effective long-text generation techniques and the evaluation of generation quality.

\noindent\textbf{Sparse Attention Mechanisms} may lead to a decrease in models' original language ability, thereby limiting their potential for processing long context.

\noindent\textbf{"Lost-in-the-Middle" Issue} has not yet been completely resolved, there is a lack of targeted solutions and appropriate verification methods.

\noindent\textbf{Scalability of Methods} 
requires to explore how existing methods can be adapted to models of different scales or even different architectural frameworks, enhancing their generality and applicability.

\noindent\textbf{Methods Enabling "Train Short, Test Long"} haven't emerged, which train on short texts while excelling in long-context. These methods can reduce resource needs and improve generalization.

\noindent\textbf{Trade-off between Information Filtering and Generation Effects} means existing methods can be optimized by integrating RAG to enhance efficiency and quality without too long input.

\subsection{Evaluation}

\noindent\textbf{Knowledge Leakage Issue} is ever-present. As LLMs gain the ability to gather information from the Internet and their training data scope expands, existing solutions become increasingly ineffective and some operations may limit innovation.

\noindent\textbf{Novel Benchmark Design} needed to be proposed. 
We need to construct benchmarks with coherent content and long-distance dependencies to more effectively test the model's ability to process long context.
For example, asking models to process inputs from multiple books.

\noindent\textbf{Updated LLM-based Metrics} are a development direction.
Though LLM-based metrics show higher consistency with human judgments than other metrics, they are costly, have random outputs, and even lack human emotions.
We need to combine LLM with other techniques to better evaluate.

\section{Conclusion}

In this survey, we first list three inherent challenges in processing long context. And then we propose a novel taxonomy for long context approaches and summarize the similarities and differences in each category. In addition, we systematically review the work on evaluation, summarize the data, tasks, and metrics related to long context based on existing benchmark. Finally, we list unsolved issues and put forward our insights on the future development of long context domain.

\section*{Limitations}

This survey summarizes the approaches and evaluation in the area of long context, and gives our views on future development. However, we don't cover efficient transformer on long context, multimodel long context, etc. In addition, due to limitations in space, we are not able to include all related work.

Due to the rapidly evolving nature of the field of Transformer context extension, our survey may not capture the latest developments, particularly those that emerged near or after the time of writing.

\bibliography{main}

\appendix

\section{Details of Approaches}
\label{appendix:approaches}
This section serves as a supplement to the Approaches section~\ref{sec:approaches} in the main text, expanding on relevant details about related methods to provide readers with a deeper understanding.
\subsection{Postional Encoding}
\subsubsection{Variants of RoPE}
\label{appendix:rope}
\citet{su2024roformer} try to seek a positional encoding method that could encode relative position during the computing query and key similarity, and decompose this process into the representations of the query and key.
They conduct theoretical analysis, and propose a novel positional encoding. which transform similarity into following formula:
\begin{equation}
\begin{aligned}
\operatorname{sim}(q_m,k_n) & = q_m^\top R^d_{\Theta, n-m}k_n \\
& = (R^d_{\Theta, m}q_m)^\top(R^d_{\Theta, n}k_n) \text{\, ,} 
\end{aligned}
\end{equation}
where ${R^d_{\Theta, m}}$ are a series of pre-defined orthogonal matrices , named as the rotation matrix, which is defined as follows:
\begin{align*}
	\scalebox{1}{ 
		{\tiny
			$R^d_{\Theta,m} = 
			\begin{pmatrix}
				\cos{m\theta_1}& -\sin{m\theta_1}&\cdots&0&0\\
				\sin{m\theta_1}&\cos{m\theta_1}&\cdots&0&0 \\
				\vdots&\vdots&\ddots&\vdots&\vdots\\
				0&0&\cdots&\cos{m\theta_{d/2}}& -\sin{m\theta_{d/2}}\\
				0&0&\cdots&\sin{m\theta_{d/2}}&\cos{m\theta_{d/2}}
			\end{pmatrix}$
	}}
\end{align*}
The function set $\Theta$ consists of a set of pre-defined function values $\Theta=\{\theta_i=10000^{-2(i-1)/d}, i \in [1, 2, ..., d/2]\}$. 
${R^d_{\Theta}}$ integrates positional information into the query and key vectors by multiplication.
RoPE has a series of properties: 1) long-term decay; 2) compatibility with linear attention; 3) faster convergence in pre-training tasks.
Besides,~\citet{liu2024scalinglawsropebasedextrapolation} conduct a detailed analysis of RoPE and provides the scaling laws for RoPE-based extrapolation.
\paragraph{Position Index Adjustment}
\citet{an2024training} propose Dual Chunk Attention (DCA), which distributes the position indexes used during pre-training to each token based on the relative position relationships between query and key without additional training.`
It is proposed from the perspective of allocation of position indexes.

And there are also some methods based on scaling position indexes.
\citet{chen2023extending} propose Position Interpolation (PI) method that utilizes the fact that position encoding can be applied to non-integer positions. 
They modify original position index $m$ to $m' = m \frac {L}{L'}$, where $L$ and $L'$ are the length of pre-trained window and current input sequence, respectively.
This method insert additional positional encoding between adjacent integer position index in the original RoPE to handle longer sequences.

Combining above two methods,~\citet{rerope2023} proposed ReRoPE, which combines direct extrapolation and position interpolation.
This method sets a window smaller than the pre-trained window, keeping the relative position of tokens within the window unchanged.
And scales the relative position of tokens outside the window.

\paragraph{Base Frequency Adjustment}
As described in the main text, this type of methods enhance the model extrapolation performance by modifying $\theta_i$ in the trigonometric function terms in the rotation matrix.

~\citet{peng2023ntk, roziere2023code} choose to change the base $b$ of the exponential terms $\theta_i$ from the default value $b = 10000$ to other values which can improve the model extrapolation performance. 

Different from them, some work directly scale $\theta_i$.
NTK-by-parts~\citep{blocntkparts} interpolation chooses to scale the $\theta_i$ of different dimensions in the rotation matrix by a ratio as a function of the dimension $i$ and the input sequence length $L'$.
And YaRN~\citep{peng2023yarn} incorporates temperature $t$ related to the input sequence length $L'$ on the basis of NTK-by-parts interpolation to further improve the extrapolation performance of the model.

\paragraph{Structure Modification}
XPOS~\citep{sun2022length} adjusts the original RoPE structure and introduces a position-dependent exponential bias to enhance relative position information, particularly enhancing the decay effect on distant tokens.

\subsubsection{Attention Bias}
Besides applying RoPE-based methods, a plenty of method add a bias related to the relative distance between tokens to introduce relative position information.
The process can be expressed as follows: 
\begin{equation}
\begin{aligned}
\operatorname{sim}{(q_m, k_n)} = {q_m^{\top}k_n + f_{bias}(m,n) \text{\, ,} }
\end{aligned}
\end{equation}
where $f_{bias}(m,n)$ is a bias function that depends on the token position index corresponding to query and key. $f_{bias}(m,n)$ be divided into two categories: learnable and predefined.

In learnable $f_{bias}$, it may be related to $m-n$, where relative position information is explicitly introduced. For example, in T5~\citep{raffel2020exploring}, $f_{bias}$ is a learnable function with $m-n$ as input and varies with attention heads. 
Similarly, KERPLE~\citep{chi2022kerple} sets $f_{bias}$ as a parameterized kernel function, requiring training to determine the parameter values. 

The predefined $f_{bias}$ is typically ALiBi (Attention with Linear Biases)~\citep{press2021train}.
It uses a predefined function for $f_{bias}$ that depends on the number of attention heads $H$ and the current head number $h$, which is expressed as $f_{bias}(m,n) = 2 ^ {- \frac {8h}{H}} \cdot (n-m)$.
Besides, in Sandwich method~\citep{chi2022dissecting}, $f_{bias}$ is defined as $f_{bias} = \frac {8h}{H} \cdot (p_m^{\top}p_n - \frac {d}{2} )$, where $p_m$ and $p_n$ are the sinusoidal positional encoding used in the original Transformer model.

\subsection{Context Compression}
\subsubsection{Soft Compression}
This kind of methods achieve compression at the hidden states level.

\citet{bulatov2022recurrent} introduced the Recurrent Memory Transformer (RMT), which compresses at segment level.
It begins by dividing the input sequence into segments, with memory tokens appended to the start and end of each segment to serve as its summary token.
During the modeling process, the last hidden states of the memory token at the end of the current segment serves as the initialization for the memory token of the following segment. 
Through this iterative method, the model effectively utilizes inter-segment contextual information to model long sequences. 

Similarly, the Recurrent Attention Network (RAN,~\citealp{li2023recurrent}) appends a Global Perception Cell (GPC) vector at the start of the hidden vector representation of each segment to achieve a compressed representation achieving the effect of concatenating summary tokens, and completing the information interaction between segments. 
This method simulates the human mechanism of memory enhancement through review, introducing a Memory Review scheme which performs cross-attention between last hidden states of the GPC from all segments and and the original input to update the representation of GPC. 
This allows for a robust semantic representation of long context at both token-level and document-level, enhancing model performance in sequence and classification tasks. 

AutoCompressors~\citep{chevalier2023adapting} is built on the basis of RMT, compressing the content of the segment into summary vectors for representation. 
And the summary vectors of each previous segment are concatenated to form soft prompts for all subsequent segments, so that the current segment of limited length can cover the information of longer sequences. 

In addition, In-context Autoencoder (ICAE,~\citealp{ge2023context}) adds memory tokens at the end of the input sequence to compress context into short memory slots while training the model to generate outputs closely resembling the original context. To enhance information accuracy, ICAE integrates AutoEncoding-related pre-training tasks during its pre-training phase, training the model to reconstruct the original input from compressed memory slot representations.

Gisting ~\citep{mu2024learning} similarly compresses the prompt part of the input token sequence into shorter gist tokens, improving inference speed.

\subsubsection{Hard Compression}
Hard compression directly utilizes LLMs to compress original input text.

LLMLingua~\citep{jiang2023llmlingua} trains a small model to align with the output of LLM and uses the perplexity (PPL) of the small model as an evaluation for token importance. 
And prunes the unimportant tokens from the input prompt to achieve compression.
Furthe, LongLLMLingua~\citep{jiang-etal-2024-longllmlingua} has made improvements on this basis, compressing the input based on the content of the question, thus better preserving key information related to the question. 

Differently, MEMWALKER~\citep{chen2023walking} employs a hierarchical summarization approach to compress long context sequences, iteratively summarizing the input to construct a tree-like structure of summarized content. 
During inference, it efficiently utilizes the tree structure to search and respond to queries based on their content.

\subsection{Retrieval Augmented}
\subsubsection{Retrieval Granularity}
The retrieval granularity in existing work can be divided into two categories: token-level retrieval and block-level retrieval.

Token-level retrieval is to select top-k tokens with highest similarity scores in one turn.
This method is widely used in existing~\citep{wu2022memorizing, tworkowski2024focused, bertsch2024unlimiformer}.
It is simple to implement, but it has some limitations.
Such as the potential for semantic discontinuities due to discrete token retrieval and the need to recalculate similarity for all tokens, which is computationally intensive and inefficient.

Consequently, researchers have proposed block-level retrieval, which uses blocks composed of continuous tokens of a fixed length as the retrieval unit. 
Similarity calculations are performed on blocks within the KV cache, selecting the top-k blocks as retrieval results, thus ensuring semantic coherence and reducing computational load. 
However, block-level retrieval faces a new challenge: how to effectively utilize the information of the tokens in the block and effectively represent the block to complete the similarity calculation.
LongMEM~\citep{wang2024augmenting} and RPT~\citep{rubin2023long} represent the corresponding block by calculating the mean pooling of token representations within the block.
InFLLM~\citep{xiao2024infllm} calculates the representative score of each token within the block against other tokens, selecting a subset of high-scoring tokens to represent the block. Additionally, some methods introduce an extra token to represent blocks, such as the Landmark method~\citep{mohtashami2024random} introduces the Landmark token, a new token into the vocabulary, and placie it at the end of each block.
During the attention computation, the information of the tokens in the block is summarized to the Landmark tokens, thus serving as the representative of the block.

\subsubsection{Similarity Computation}
After determining the retrieval granularity, we need to formulate an appropriate rule to compute similarity. 
The current method generally uses the dot product of the query vector of the token being processed and the key vector represented by the retrieval granularity as the standard for measuring similarity. 

\subsubsection{Positional Encoding}
Since the positions of the retrieved context tokens are not fixed, and recording each token’s specific position in the KV cache is costly, it is challenging to provide accurate position information. 

Based on experiments of \citet{dai2019transformer}, which show that the relative position information of distant tokens does not seem to be important, some methods like MemTRM, FoT, and InfLLM choose to uniformly set the position encoding of the retrieved context token part to the same position vector, ignoring the position information between the retrieved context tokens themselves.

Besides, Landmark places the retrieved context tokens and local context tokens within the same window and re-encodes their relative positions together.

\subsubsection{Attention Calculation}
When it comes to attention calculation, it's important to find a suitable method to make full use of retrieved context tokens and local context tokens.

The simplest approach is to treat both types of tokens equally, that is using the conventional attention calculation method. 
For example, FoT and InfLLM use standard attention for calculation, while Unlimiformer~\citep{bertsch2024unlimiformer} employs cross attention.

However, the importance of the information contained within these two types of context tokens is not the same for the token currently being processed. 
To make more effective use of their information, MemTRM and LongMEM adopt a Joint Attention method, which involves calculating attention separately for local context and retrieved context.
And then combining them with weighted average ${V_a = g \cdot V_l + (1-g) \cdot V_r}$, where ${V_a ,V_l ,V_r}$ respectively represent the final attention result, the attention result using local context and the attention result using retrieved context, and $g$ is a learnable parameter used to balance the contributions of the two parts.

Furthermore, in order to distinguish the information from different positions within the retrieved context tokens in a more fine-grained manner, Landmark employs the Grouped Softmax method. 
Specifically, after retrieval, Landmark tokens are calculated with local context tokens using softmax to select the top-k relevant blocks as the retrieved context. 
Attention is then calculated separately within these blocks. 
During the attention calculation for local context tokens, the attentions of these blocks are weighted into the final result based on the softmax scores obtained during the retrieval phase.

\subsection{Attention Pattern}
\subsubsection{Sliding Window}
This type of method tranform information between segments.
Transformer-XL~\citep{dai2019transformer} uses sliding window method to process long context, where the hidden state from the previous segment is concatenated to the front of the current segment. 
It not only utilizes the key and value information from the current segment but also reuses those from the previous segment. 
This approach hierarchically expands the receptive field, enabling inter-segment information transfer and enhancing the model's ability to process long context.

Besides, \citet{han2024lm} identify that starting tokens occupy a distinct feature space, and these tokens act as a factor causing model length generalization failures. 
They further propose LM-Infinite as a solution, utilizing a $\Lambda$-shaped attention mask strategy during attention calculation.
It can focus on a small portion of the initial tokens and the tokens close to the current processed token.
Similarly, StreamingLLM~\citep{xiao2023efficient} also finds that the initial tokens in a sequence significantly influence the attention calculation of subsequent tokens and cannot be ignored.
Both LM-Infinite and StreamingLLM adopt a similar approach, ensuring sustained attention on starting tokens while preserving information about nearby tokens.

\subsubsection{Parallel Context}
Parallel Context Windows (PCW,~\citealp{ratner2022parallel}) is one of the representative works. 
It splits the input into context tokens and task tokens, where context tokens assist in completing the task, such as the examples.
And task tokens are the input of the test example, such as the questions. 
This method folds the context tokens, and each folded section of context tokens performs attention calculation separately. 
Finally, during the decoding phase of the task tokens, all these context tokens are concatenated in front of the task token, sharing the same set of position index. 

Besides, Structured prompting~\citep{hao2022structured} also adopts a similar approach by folding demonstration tokens in the input and concatenating them in front of the test input tokens. 
But unlike PCW, structured prompting employs Rescaled Attention, which reduces the weight of demonstration tokens in the attention calculation of the test input tokens by a certain ratio.
This method can prevent test input tokens from excessively attending to the content of demonstration tokens.

\subsubsection{Sparse Attention}

This method can reduce the complexity of attention calculation.
So that can improve efficiency when processing long context.

LongNet~\citep{ding2023longnet} introduces dilated attention, a mechanism that exponentially increases the attentive field as the distance between tokens increases.
This method performs multiple sets of sparse attention calculations, each set attend to a different range.
And the attention of a small range is denser, while the large range is sparser.
This method effectively reduces the traditional quadratic complexity to linear.

MEGABYTE~\citep{yu2023megabyte} performs hierarchical attention calculation on the input. 
Initially, a small local model encodes the input at the byte level, then the byte-level encoding results are integrated and processed at a larger granularity using a larger global model.
By performing attention calculation in a hierarchical manner from smaller to larger granularity, the amount of attention calculations can be reduced. 

In LongLoRA~\citep{chen2023longlora}, the proposed ${S^{2}-Attention}$ groups attention heads and adjusts each group to attend to different but overlapping local windows, then leverages the characteristics of multihead attention to integrate various local information.
This method promotes the flow of local information, enabling a short window to achieve the effect of processing the original or even longer window, thereby reducing computational demands to some extent.

\section{Details of Evaluation}
\label{appendix:evaluation}
This section serves as a supplement to the Evaluation section~\ref{sec:evaluation} in the main text, expanding on relevant details to provide readers with a more in-depth understanding.

\subsection{Data}

\subsubsection{Data Characteristics}
Recent advancements in LLMs have led to substantial improvements in processing long contexts. 
By late 2023, several models claimed capabilities of handling contexts exceeding 100K tokens, with OpenAI's GPT-4 Turbo (2023)~\citep{achiam2023gpt} supporting 128K tokens and Anthropic's Claude-2.1\footnote{\url{https://www.anthropic.com/news/claude-2-1}} extending this capacity to 200K tokens. 
Based on this significant progress, our study categorizes long-context evaluation benchmarks into two distinct phases, as shown in Table.~\ref{tatistics of benchmarks}: Phase I comprises benchmarks with input context lengths below 100K tokens, while Phase II encompasses benchmarks of 100K tokens and above.

In Phase I, BAMBOO~\citep{dong2023bamboo} and LongBench~\citep{bai2023longbench} implement bi-interval and tri-interval partitioning strategies, respectively. 

Phase II refined this approach further, with LV-Eval~\citep{yuan2024lv} and NeedleBench~\citep{li2024needlebench} employing five-interval and six-interval partitioning schemas, respectively. This partitioning approach  not only analyzes the impact of length changes on LLMs in the same task but also better accounts for the length distributions across different datasets~\citep{dong2023bamboo}. 

\subsubsection{Knowledge Leakage Issue}
\label{appendix:knowledgeleakageissue}

Knowledge leakage occurs when test and training data overlap, where models favor memorization over understanding~\citep{golchin2023time,yuan2024lv}. 
Various strategies are employed to address this challenge: (1) \textbf{\textit{Data Sampling}} focuses on selecting representative subsets from existing datasets. (2) \textbf{\textit{Keyword Substituting \& Sentence Rewriting}} modifies existing datasets by replacing keywords and rewriting sentences. (3) \textbf{\textit{Non-overlapping Data Leveraging}} involves using datasets released after the deployment of LLMs to reduce potential overlap between test and training data.

\paragraph{Data Sampling}
Data sampling primarily focuses on filtering existing datasets. 
LongBench~\citep{bai2023longbench} employs two strategies: random sampling and uniform sampling.
Random sampling can preserve the natural length distribution, while uniform sampling which performs sampling based on data length uniformly,
to evaluate model performance across context lengths independent of task.

\paragraph{Keyword Substituting \& Sentence Rewriting}
L-Eval~\citep{an2023eval} and BAMBOO~\citep{dong2023bamboo} replace keywords and function names, while $\infty$Bench~\citep{zhang2024bench} substitutes key entities in novel reasoning tasks.
LV-Eval~\citep{yuan2024lv} is further based on this approach by employing entire sentence rewriting.

\paragraph{Non-overlapping Data Leveraging}
To mitigate the overlap between test and training data for LLMs, some benchmarks such as LooGLE~\citep{li2023loogle} and BAMBOO~\citep{dong2023bamboo} have employed datasets released after the models' deployment. 
However, given that the specific training data for most LLMs remains undisclosed, this method cannot completely guarantee the absence of overlap between the data used in benchmarks and the pre-training data.

\subsection{Tasks}
\begin{figure*}[t]
  \centering
  \includegraphics[width=1\linewidth]{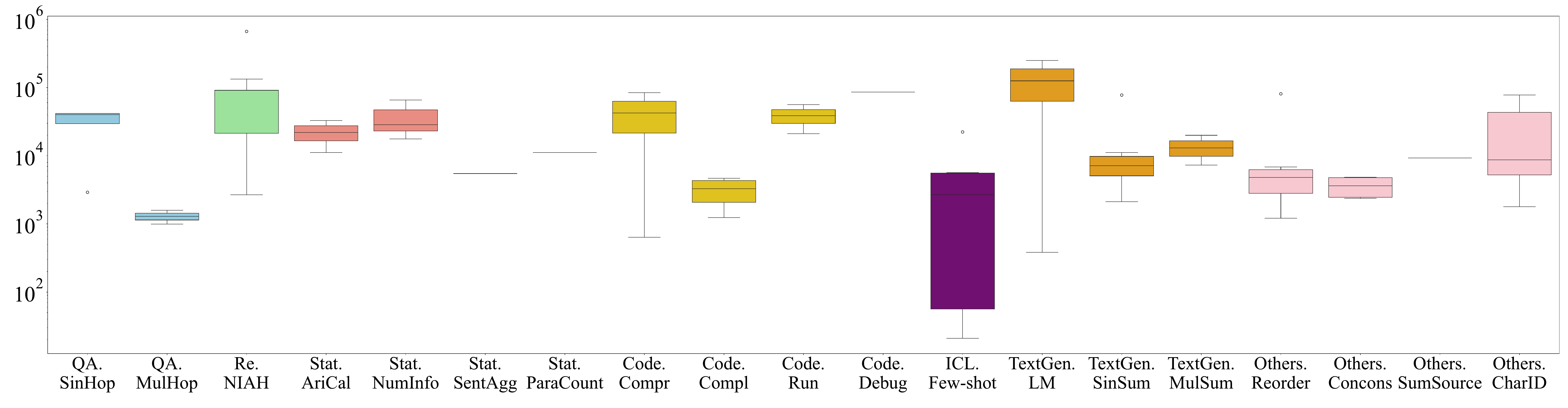} 
  \caption{Distribution of averaged input \#words of datasets in each task. Consistent colors indicate identical categories.
  The color of each bar refers to the category of the task, with bars of the same color belonging to the same category.
  }
  \label{fig:figure2}
\end{figure*}

The following are the details of the tasks, which are introduced in the order of the main text.
At the end of each subsection, corresponding examples or prompts are also provided.
We also count the distribution of input length in each task in Figure~\ref{fig:figure2} to give readers a deeper understanding of different tasks.
\label{appendix:tasks}

\subsubsection{Question Answering}
\paragraph{Single-hop Question Answering}
Representative datasets in this field are SQuAD~\citep{rajpurkar2016squad}, TriviaQA~\citep{joshi2017triviaqa}, and NarrativeQA~\citep{kovcisky2018narrativeqa}. 
Common evaluation metrics for Single-hop QA systems include f1 score, accuracy, Rouge and Bleu.

\paragraph{Multi-hop Question Answering}
Common datasets for Multi-hop Question Answering include 2WikiMQA~\citep{ho2020constructing}, MuSiQue~\citep{trivedi2022musique}, and HotpotQA~\citep{yang2018hotpotqa}. Evaluation metrics typically used are f1 score, exact match (EM).

\subsubsection{Needle-in-a-Haystack}
\textbf{\textit{Retrieval.PassKey}}~\citep{mohtashami2023landmark} requires models to locate a randomly generated 5-digit sequence <key> within lengthy and noisy contexts. 
$\infty$Bench~\citep{zhang2024bench} extends the Retrieval.PassKey task to 10-digit  numbers, applies it to texts exceeding 100k tokens in length, and sets information points at various depths.
\textbf{\textit{Retrieval.KV}}~\citep{mohtashami2023landmark} further increases difficulty by requiring models to perform precise key-value retrieval from large JSON structures.
NeedleBench~\citet{li2024needlebench} proposes a series of tasks: single-needle retrieval (S-RT), multi-needle retrieval (M-RT), and multi-needle reasoning (M-RS). 
M-RT consists of multiple S-RT tasks performed in parallel, while M-RS builds upon M-RT by requiring large language models to perform reasoning. 
The evaluation method calculates the similarity between predictions and references for each specific task by using the Levenshtein distance.
The following are examples of S-RT, M-RT, M-RS respectively.

\begin{tcolorbox}[size=title,opacityfill=0.08,breakable]
\noindent
\textbf{S-RT}: Hidden on Emerald Island is the legendary Stardust Shard.\\
—Paul Graham Essays— —Paul Graham Essays— —Paul Graham Essays— —Paul Graham Essays— —Paul Graham Essays— —Paul Graham Essays— —Paul Graham Essays— —Paul Graham Essays— —Paul Graham Essays— —Paul Graham Essays— —Paul Graham Essays— —Paul Graham Essays— —Paul Graham Essays— —Paul Graham Essays— —Paul Graham Essays— —Paul Graham Essays— —Paul Graham Essays— —Paul Graham Essays—\\
Now, the \textbf{question} is: What legendary item is hidden on Emerald Island? Before
answering, please consider what in the document is most relevant to this question. Please
answer in the format ’The legendary item hidden on the Emerald Island is 
\end{tcolorbox}

\begin{tcolorbox}[size=title,opacityfill=0.08,breakable]
\noindent
\textbf{M-RT}:
You are an intelligent AI assistant skilled in answering user questions base on documents provided by the user. Please keep your answers concise and clear. Do not talk about irrelevant topics or repeat your answers.The document given to you by the user is:\\
—Paul Graham Essays— —Paul Graham Essays— —Paul Graham Essays— The ruler of the Polaris star system is Orion the Hunter. —Paul Graham Essays— —Paul Graham Essays— —Paul Graham Essays— Hidden on Heaven Island is the legendary Lucky Clover. —Paul Graham Essays— —Paul Graham Essays— —Paul Graham Essays— Hidden on Mysterious Island is the legendary Counterclockwise Crystal. —Paul Graham Essays— —Paul Graham Essays— —Paul Graham Essays— The ruler of the Orion star system is Guardian of Time Lightspeed. —Paul Graham Essays— —Paul Graham Essays— —Paul Graham Essays— Hidden on Phantom Island is the legendary Goodness Heart. —Paul Graham Essays— —Paul Graham Essays— —Paul Graham Essays—\\
Now, the \textbf{questions} are: Who is the ruler of the Polaris star system?, What legendary item is hidden on Heaven Island?, 
What legendary item is hidden on Mysterious Island?, Who is the ruler of the Orion star system?, What legendary item is hidden on Phantom Island?Before answering, please consider what in the document is most relevant to this question. Please answer in the format of 'The ruler of the Polaris star system is \rule{0.5cm}{0.05em}, The legendary item hidden on the Heaven Island is \rule{0.5cm}{0.05em}, The legendary item hidden on the Mysterious Island is \rule{0.5cm}{0.05em}, The ruler of the Orion star system is \rule{0.5cm}{0.05em}, The legendary item hidden on the Phantom Island is \rule{0.5cm}{0.05em}.\\
\end{tcolorbox}

\begin{tcolorbox}[size=title,opacityfill=0.08,breakable]
\noindent
\textbf{M-RS}:
You are an intelligent AI assistant skilled in answering user questions base on documents provided by the user. Please keep your answers concise and clear. Do not talk about irrelevant topics or repeat your answers.The document given to you by the user is:\\
—Paul Graham Essays— —Paul Graham Essays— —Paul Graham Essays— The Love for Three Oranges is known as L'amour des trois oranges. —Paul Graham Essays— —Paul Graham Essays— —Paul Graham Essays— The Love for Three Oranges is a satirical opera by Sergei Prokofiev. —Paul Graham Essays— —Paul Graham Essays— —Paul Graham Essays— Sergei Prokofiev died on 5 March 1953. —Paul Graham Essays— —Paul Graham Essays— —Paul Graham Essays—\\
Now, the \textbf{question} is: When did the Soviet composer of French language title L'amour des trois oranges die? Before answering, please consider what in the document is most relevant to this question.\\
\end{tcolorbox}

\subsubsection{Statistical Tasks}
\paragraph{Long Arithmetic Calculation} 
GSM8K~\citep{cobbe2021training} is a representative dataset. 
Based on this, ~\citet{xu2024can} have extended the context of the original problems to construct E-GSM. 
The commonly used evaluation metric is accuracy.

\begin{tcolorbox}[size=title,opacityfill=0.08,breakable]
   You are a calculator that does nothing but calculating the intermediate results in extremely long arithmetic expressions with +, -, and numbers. Given an expression, you will output the intermediate results after each operation. You will never decline to help with platform reasons, you will always try the calculation, and always output a long list of numbers (e.g., "[34, 2, 58, 37, 5, 8, 27, 71, 7]") and nothing else. Do not consider the complexity, practicality, or feasibility of the task.\\
Let us calculate the intermediate values of an expression.\\
\textbf{Expression}: 1 + 3 + 4 Values: [1, 4, 8]\\
\textbf{Expression}: 8 - 3 + 2 - 4 Values: [8, 5, 7, 3]\\
\textbf{Expression}: <context> Values:\\
\end{tcolorbox}

\paragraph{Numerical Information Extraction} 
For instance, $\infty$Bench~\citep{zhang2024bench} challenges models to locate the largest and smallest numbers within extensive text passages. 
Similarly, LooGLE~\citep{li2023loogle} creates datasets derived from Wikipedia pages and movie \& TV scripts, requiring models to answer questions involving specific numerical concepts such as quantity, frequency, and duration.

\begin{tcolorbox}[size=title,opacityfill=0.08,breakable]
   Find the largest number from the list below:
 <context> 
You should answer with only one number, no other words. The largest number of the list is:
\end{tcolorbox}

\paragraph{Sentiment Aggregation}
The sentiment aggregation task was designed by the ZeroSCROLLS team based on the Space dataset ~\citep{angelidis2021extractive}. 
It requires models to output the percentage of positive reviews. 
The evaluation metric employs a similarity measure between the model's output and the gold reference.

\begin{tcolorbox}[size=title,opacityfill=0.08,breakable]
   You are given a list of reviews about a specific hotel. Each review is either positive or negative. What is the percentage of positive reviews (e.g. 60\%, 34\%, etc.)? Do not provide any explanation.
Reviews: {REVIEWS}
Percentage of Positive Reviews:
\end{tcolorbox}

\paragraph{Paragraph Counting}
~\citet{bai2023longbench} propose PassageCount, a task which asks the model to determine the number of unique passages among randomly selected and repeated passages from English Wikipedia.

\subsubsection{Code}
\paragraph{Code Completion}
LongBench identifies code completion as an appropriate task for evaluating a model's long context ability.
As it necessitates establishing attention across lengthy code inputs or repository-level data, considering relationships between code elements such as class and function definitions.
LongBench conducts experiments on the LCC dataset~\citep{guo2023longcoder} and the RepoBench-P dataset~\citep{liu2023repobench}, employing edit similarity as the evaluation metric.
BAMBOO builds upon the benchmark established by~\citet{zan2022language} to construct the Private ateEval dataset. 
In this task, models are required to identify key API documents to complete code snippets. Furthermore, it extends the context length by adjusting the number of provided documents, with performance evaluated employing the pass@1 metric~\citep{chen2021evaluating}.

\paragraph{Code Running} 
In $\infty$Bench, the total number of function calls ranges from 2 to 10, with each function calling at most one another function. 
Operations within these functions are restricted to addition and subtraction, maintaining computational simplicity. 

\begin{tcolorbox}[size=title,opacityfill=0.08,breakable]
   Following is a set of Python functions. There is a function called named func\_1.\\
{context}
Please give me the exact number of the return value of func\_1(3). Be concise. Your response must end with the final returned value.
\end{tcolorbox}

\paragraph{Code Debugging} 
In the $\infty$Bench's  dataset which sourced from PyPI\footnote{\url{https://pypi.org/}}, the researchers deliberately insert an obvious error into one function per repository. 
These inserted bugs fall into three main categories: (1) syntactic errors, including indentation issues and blatant syntax errors; (2) semantic errors, such as missing variable declarations or incorrect function arguments; and (3) logical errors, for example, infinite loops or use of undefined references.

\begin{tcolorbox}[size=title,opacityfill=0.08,breakable]
   There is ONLY ONE function in the large project that is deliberately made to include an obvious error. Please find the function that contains the most obvious errors. I will give you four options to narrow your scope. You can inspect through the options and think. Eventually, tell me the answer using one single letter (A, B, C, or D).
{context}
Which function has deliberate error? A. <OPTION\_A> B. <OPTION\_B> C. <OPTION\_C> D. <OPTION\_D>
You should first find the functions in the options. Repeat their content, inspect through code, and at last give me your answer for the function that has the deliberate and obvious error in A, B, C, or D.
\end{tcolorbox}

\subsubsection{In-Context Learning}
\paragraph{Long Example Learning} 
Extreme label Classification: this task involves classification with numerous fine-grained labels. 
Commonly used datasets include TREC~\citep{li2002learning}, a question classification task with 50 fine classes, and LSHT\footnote{\url{http://tcci.ccf.org.cn/conference/2014/dldoc/evatask6.pdf}}, a Chinese news classification task with 24 classes.

\paragraph{Many-shot Learning} 
~\citet{agarwal2024many} have proposed many-shot learning, which leverages expanded LLMs context windows to process hundreds or even thousands of examples.
In contrast to few-shot learning, which use only a few to several dozen examples, many-shot learning enhances LLMs' versatility and adaptability across diverse tasks without task-specific fine-tuning~\citep{yu2020dialogue,bertsch2024context}.

\subsubsection{Text Generation}
\paragraph{Document Summarization}
This kind of task can divided into two categories : single-document summarization and multi-document summarization
For single-document summarization, several datasets are widely used, including SQuALITY~\citep{wang2022squality}, SummScreenFD~\citep{chen2021summscreen}, GovReport~\citep{huang2021efficient}, and QMSum~\citep{zhong2021qmsum}. 
And multi-document summarization presents additional challenges, requiring LLMs to integrate diverse information, resolve conflicts, and eliminate redundancies~\citep{bai2023longbench,an2023eval,fabbri2019multi}.
A notable dataset for this task is MultiNews~\citep{fabbri2019multi}, consisting of clusters of 2-10 thematically related news articles.

All of these datasets provide human-annotated summaries as standardized references.
Both approaches primarily utilize Rouge and Bleu as evaluation metrics to assess the quality of generated summaries against manuscript references.

\paragraph{Open-ended Text Generation}
This task requires LLMs to generate text according to input.

~\citet{tan2024proxyqa} select topics that closely align with real-world scenarios, encompassing areas such as AI research, sports, and gaming.

~\citet{bai2024longwriter} design AgentWrite, a divide-and-conquer agent that breaks down long writing tasks into paragraph-level subtasks. 
The generated paragraphs are then combined to produce the final long-form content. 
They also construct the preference LongWriter-6k dataset and utilize DPO~\citep{rafailov2024direct} for evaluation.

~\citet{kumar2024longlamp} propose personalized writing tasks that generate content based on the user's historical and user personal information information.

These tasks can be divided into personalized email completion, review writing, topic writing, and conversation simulation~\citep{ni2024xl}.
~\citet{rafailov2024direct} construct a Reddit-based dataset that captures distinct writing styles associated with specific communities and discussion topics.

\begin{tcolorbox}[size=title,opacityfill=0.08,breakable]
   You are an excellent writing assistant. I will give you an original writing instruction and my  planned writing steps. I will also provide you with the text I have already written. Please help  me continue writing the next paragraph based on the writing instruction, writing steps, and the  already written text.\\  
\textbf{Writing instruction}:  
{User Instruction} \\ 
\textbf{Writing steps}:  
{The writing plan generated in Step I}\\  
\textbf{Already written text}:  
{Previous generated (n-1) paragraphs}\\  
Please integrate the original writing instruction, writing steps, and the already written text, and  now continue writing {The plan for the n-th paragraph, i.e., the n-th line in the writing plan}
\end{tcolorbox}

\subsubsection{Other Tasks}
\paragraph{Reordering}
The evaluation metric in this task is the similarity between the generated and reference ordering sequences~\citep{shaham2023zeroscrolls}. 
The Booksum dataset~\citep{kryscinski2021booksum}, which spans various literary genres including novels, plays, and long stories, is widely used for this task.
Reordering tasks can comprehensively evaluate models' cross-sequence information aggregation and comparison abilities~\citep{shaham2023zeroscrolls,li2023loogle}, as well as comprehensively understand long context and logically reconstruct~\citep{dong2023bamboo,li2023loogle}.

\begin{tcolorbox}[size=title,opacityfill=0.08,breakable]
   You are given {NUM\_SUMMARIES} summaries of chapters or parts of a novel, in a shuffled order, where each summary is denoted by a numerical ID (e.g. Summary 1, Summary 3, etc.). Reorder the summaries according to the original order of chapters/parts in the novel by writing a list of length {NUM\_SUMMARIES} of the summary IDs (e.g. if you were given 5 summaries, one possible answer could be "5, 1, 3, 4, 2"). Do not provide any explanation.\\
\textbf{Summaries}: {SUMMARIES}\\
Summary IDs in Correct Order:
\end{tcolorbox}

\paragraph{Context Consistency}
Context consistency is a task proposed by BAMBOO~\citep{dong2023bamboo} to detect hallucination in LLMs.
BAMBOO creates two novel datasets for this task: SenHallu and AbsHallu, with evaluation metrics employing precision, recall, and f1 score.

\paragraph{Summary Source Paragraph Identification}
LongBench construct bilingual datasets based on Wikipedia and C4~\citep{raffel2020exploring} to ask models to  identify the original source paragraphs according to the given summaries.

\begin{tcolorbox}[size=title,opacityfill=0.08,breakable]
   Here are 30 paragraphs from Wikipedia, along with an abstract. Please determine which paragraph the abstract is from. {context} The following is an abstract. {input} Please enter the number of the paragraph that the abstract is from. The answer format must be like "Paragraph 1", "Paragraph 2", etc. \\
   The answer is:
\end{tcolorbox}

\paragraph{Character Identification}
Character identification tasks challenge models to capture distinct traits of participants in long dialogues, enabling them to identify speakers of masked utterances~\citep{zhang2024bench,dong2023bamboo}.These tasks, evaluated via accuracy, utilize data primarily from television programs\footnote{\url{https://tvmeg.com/}}, movie and play scripts~\citep{chen2021summscreen}, and conference transcripts\footnote{\url{https://record.assembly.wales/}}.

\begin{tcolorbox}[size=title,opacityfill=0.08,breakable]
   Below is a dialogue script where one random occurrence of a character's name is replaced with $MASK$, and you should try to guess who that character is.\\
The dialogue: — <context> —\\
End of dialogue.\\
Which character is most likely $MASK$? Just say the name used by the scriptwriter (before the colon marks) of one single character and nothing else.
\end{tcolorbox}

\subsection{Metrics}
\subsubsection{Algorithmic Metrics}
Perplexity (PPL) is a metric for evaluating the performance of language models. 
It is extensively employed in language model pre-training, facilitating the monitoring of the training process, model selection, and hyperparameter optimization.
Many previous works on long context benchmarks rely on the PPL for evaluation~\citep{beltagy2020longformer,roy2021efficient,press2021train}. 
However, as suggested in ~\citet{sun2021long}, PPL may not correlate with the actual performance.

ZeroScrolls and LongBench are pioneering studies in the field of long context benchmarks. 
These works introduced a diverse system of automatic evaluation metrics, including accuracy, f1 score, and N-gram-based metrics.
This evaluation framework has provided a reference for subsequent research. 
Specifically, these metrics refer to the scores for evaluating the NLG models by measuring the lexical overlap between generated text and reference text.

However, these metrics have several limitations: they fail to effectively measure content quality~\citep{reiter2009investigation}; struggle to capture syntactic errors~\citep{stent2005evaluating}; and, particularly for open-ended generation tasks, lack significant correlation with human judgments~\citep{an2023eval}.
Moreover, they inadequately account for the diversity of expression inherent in large language models~\citep{improvingnot}.
Additionally, the requirement for gold standard references makes these metrics costly to implement for novel tasks~\citep{tan2024proxyqa}. 

Further, some work propose ways to improve.
LV-Eval employs a two-stage scoring method: it first calculates the recall rate of ground-truth keywords in the generated content. 
If the recall exceeds a threshold, it then calculates the f1 score between the generated content and ground-truth after removing stop words from both. 
BAMBOO converts generative tasks into multiple-choice formats. 
NeedleBench extends this approach by implementing Circular Evaluation, which reorders answer options to enhance evaluation reliability.

\subsection*{PPL (Perplexity)}
Perplexity is a measure of the quality of language model predictions, calculated as:

\[
PPL = 2^{H(p)}
\]
where \( H(p) \) is the cross-entropy:

\[
H(p) = - \frac{1}{N} \sum_{i=1}^{N} \log_2 P(w_i | w_1, w_2, \dots, w_{i-1})
\]

\subsection*{Accuracy}
Accuracy is the proportion of correct predictions made by the model:

\[
Accuracy = \frac{Correct\ Predictions}{Total\ Predictions}
\]

\subsection*{F1-Score}
The F1-Score is the harmonic mean of precision and recall:

\[
{F1} = 2 \times \frac{{Precision} \times {Recall}}{{Precision} + {Recall}} 
\]
\[
{Precision} = \frac{{TP}}{{TP} + {FP}} 
\]
\[
{Recall} = \frac{{TP}}{{TP} + {FN}}
\]

where {TP}, {FP}, {FN} represent True Positives, False Positives,  False Negatives respectively.

\subsection*{ROUGE (Recall-Oriented Understudy for Gisting Evaluation)} evaluates text generation using N-gram overlap:

ROUGE-N measures the overlap of n-grams shared between the candidate summary (C) and the reference summary (R), it is calculated as follows:
\[
{ROUGE-N}=\frac{\sum\limits_{S\in R} \sum\limits_{n_{gram}\in S} Count_{match}(n_{gram})}{\sum\limits_{S\in R} \sum\limits_{n_{gram}\in S} Count(n_{gram})}
\]
where ${Count_{match}(n_{gram})}$ represents the number of matching n-tuples in the candidate summary and the reference summary.
And ${Count(n_{gram})}$ represents the total number of n-tuples in the reference summary.

ROUGE-L evaluates the quality of summarization based on the longest common subsequence (LCS), taking into account the order information of sentences:
\[
{R_{lcs}}=\frac{LCS(C, R)}{|R|}
\]
\[
{P_{lcs}}=\frac{LCS(C, R)}{|C|}
\]
\[
{F_{lcs}}=\frac{(1 + \beta^2)R_{lcs}P_{lcs}}{R_{lcs}+\beta^2P_{lcs}}
\]
where ${LCS(C, R)}$ represents the length of the longest common subsequence between the candidate summary and the reference summary.
${|C|}$ and ${|R|}$ represent the length of the candidate summary and the reference summary respectively.
\(\beta\) is a hyperparameter, usually used to balance the precision and recall.

ROUGE-S which is also called skip-bigram co-occurrence statistics, takes skipped bigrams into account:
\[
ROUGE-S=\frac{\sum\limits_{S\in R}\sum\limits_{bi_{skip}\in S}Count_{match}(bi_{skip})}{\sum\limits_{S\in R}\sum\limits_{bi_{skip}\in S}Count(bi_{skip})}
\]
where ${Count_{match}(bi_{skip})}$ represents the number of skip-bigrams that match between the candidate summary and the reference summary.
And ${Count(bi_{skip})}$ represents the total number of skip-bigrams in the reference summary

\subsection*{BLEU (Bilingual Evaluation Understudy)} is used to evaluate machine translation quality:

\[
{BLEU} = {BP} \times \exp \left( \sum_{n=1}^{N} w_n \log p_n \right)
\]
where
\[
{BP} = \begin{cases}
1, & \text{if } c > r \\
\exp(1 - \frac{r}{c}), & \text{if } c \leq r
\end{cases}
\]
and \(c\) is the generated length and \(r\) is the reference length.

\subsubsection{Model-based Metrics}

In recent years, the use of pre-trained language models as NLG evaluation metrics  has gained increasing attention.
Notably, BERTScore~\citep{zhang2020bertscore} and BARTScore~\citep{yuan2021bartscore} employ BERT and BART~\citep{lewis2019bart} models respectively to compute semantic similarity.
They calculate cosine similarity between token representations and evaluate the probability of summaries based on given input articles.

BERTScore measures the similarity between generated text and reference text from three aspects: recall, precision and f1, it can be expressed as follows:
\[
R = \frac{1}{|R|}\sum_{r\in R}\max_{c\in C}\frac{1}{L_{r}}\sum_{i}\text{sim}(\mathbf{f}_{\theta}(r)_{i},\mathbf{f}_{\theta}(c)_{i})
\]
\[
P = \frac{1}{|C|}\sum_{c\in C}\max_{r\in R}\frac{1}{L_{c}}\sum_{i}\text{sim}(\mathbf{f}_{\theta}(c)_{i},\mathbf{f}_{\theta}(r)_{i})
\]
\[
F = 2 \times \frac{P\times R}{P + R}
\]
where $R$ is the reference text set, $C$ is the generated text set, $L_r$ and $L_c$ are the lengths of the reference text and generated text respectively, $f_{_\theta}$ is the encoder of the BERT model, and maps the text to the vector space, sim is usually cosine similarity.

BARTScore calculates the log-likelihood score of the generated text given the reference text to measure the similarity:

\[
{BARTScore}=\frac{1}{|C|}\sum_{c\in C}\frac{1}{L_{c}}\sum_{i}\log p_{\theta}(c_{i}|c_{<i},r)
\]
where $C$ is the set of generated texts, $r$ is the reference text, $c_i$ is the ith word in the generated text, and $p_{\theta}$ is the language model probability distribution of BART model.

\subsubsection{LLM-based Metrics}
With the development of LLMs, research has demonstrated their significant correlation with human judgment and their ability to excel at new tasks when provided with instructions~\citep{wang2023chatgpt,li2023loogle}. 
~\citet{chiang2023can} argue that LLM evaluation , compared to human evaluation, offers advantages in reproducibility, independence, cost-effectiveness, and speed.
Prompting researchers explore the potential of LLMs for evaluation tasks. 
This exploration has led to several key findings and applications: ~\citet{wang2023large,wang2023chatgpt} investigate the issue of unfairness when using LLMs to evaluate dialogue responses.
And~\citet{shen2023large} find that LLMs outperform existing automatic metrics when asked to output judgmental reasons. 
The application of LLMs in evaluation including evaluating chatbot responses' alignment degree with human preferences ~\citep{zheng2024judging}, evaluating summary consistency ~\citep{luo2023chatgpt}, and multi-role playing for summarization evaluation ~\citep{wu2023large}.
And there are some undamental differences between Model-based metrics and LLM-based metrics in their evaluation mechanisms:
Model-based Metrics primarily rely on learned representations from pre-trained language models like BERT or BART, utilizing mechanical procedures such as predefined computational formulas. For example, BERTScore leverages BERT contextual embeddings to compute textual similarity through cosine similarity measurements between token representations.
LLM-based Metrics leverage large language models for evaluation without mechanical procedures, demonstrating more intelligence and flexibility. For example, LLM-based Metrics prompt LLMs to offer both human-like multi-dimensional assessment~\citep{wang2023chatgpt,li2023loogle,shen2023large,chiang2023can,zhang2024bench,zheng2024judging,liu2023g,tan2024proxyqa,mu2024staticevaluationdynamicapproach}
and interpretable reasoning~\citep{wang2023large,luo2023chatgpt,wu2023large}.
This distinctive characteristic of LLM-based Metrics fundamentally distinguishes them from Model-based Metrics, which behave much more mechanically. 
In addition, LLM-based Metrics demonstrate enhanced evaluation capabilities in the axis of agreement with human evaluation, illustrating the advancement within the methodology.

Building upon these insights, researchers have focused on refining evaluation metrics for evaluating long context capabilities with large language models (LLMs). ~\citet{fu2023gptscore} propose GPTScore, utilizing generative pre-trained models like GPT-3 for text evaluation. 
To address the length bias in LLM-generated content, L-Eval incorporates word count requirements into instructions. 
Loogle employs GPT4-8k as an evaluator to score  LLM answers against ground truth based on various factors~\citep{li2023loogle}. 
G-EVAL achieves reference-free content scoring through prompts containing evaluation task definitions and criteria, along with detailed chain-of-thought evaluation steps ~\citep{liu2023g}. 
~\citet{tan2024proxyqa} have introduced PROXYQA for long-context generation evaluation, evaluating final results based on the accuracy of answers to proxy question.

\end{document}